%% file: main.tex
\documentclass[11pt,a4paper]{article}
\usepackage[hyperref]{acl2017}
\usepackage{times}
\usepackage{latexsym}
\usepackage{makecell}

\usepackage{amsmath}
\usepackage{algorithm}
\usepackage{algpseudocode}

\usepackage{url}

\usepackage{graphicx}
\newcounter{tbsnr}
\newenvironment{tbs}
{\addtocounter{tbsnr}{1}\par\bigskip\noindent\fbox{\thetbsnr}
\hspace*{\fill}\begin{minipage}{7cm}\tt}
{\end{minipage}\hspace*{\fill}\bigskip}

\newcommand{\cut}[1]{}

\aclfinalcopy 
\def\aclpaperid{481} 

\title{FOIL it! Find One mismatch between Image and Language caption}

\author{Ravi Shekhar, Sandro Pezzelle, Yauhen Klimovich,\\ \textbf{Aur\'{e}lie
  Herbelot, Moin Nabi,  Enver Sangineto, Raffaella Bernardi}\\
University of Trento \\
{\tt \{firstname.lastname\}@unitn.it} 
}

\date{}

\begin{document}
\maketitle
\begin{abstract}

  In this paper, we aim to understand whether current language and
  vision (LaVi) models truly grasp the interaction between the two
  modalities. To this end, we propose an extension of the MS-COCO
  dataset, FOIL-COCO, which associates images with both correct and
  `foil' captions, that is, descriptions of the image that are highly
  similar to the original ones, but contain one single mistake (`foil
  word'). We show that current LaVi models fall into the traps of
  this data and perform badly on three tasks: a) caption
  classification (correct vs.\ foil); b) foil word detection; c) foil
  word correction. Humans, in contrast, have near-perfect performance
  on those tasks. We demonstrate that merely utilising language cues is not
  enough to model FOIL-COCO and that it challenges the
  state-of-the-art by requiring a fine-grained understanding of the
  relation between text and image.

\end{abstract}

\input{introduction}
\input{related_work}
\input{datasets}
\input{experiments}
\input{analysis}
\input{conclusions}

\section*{Acknowledgments}
We are greatful to the Erasmus Mundus European Master in Language and
Communication Technologies (EM LCT) for the scholarship provided to
the third author.  Moreover, we gratefully acknowledge the support of NVIDIA
Corporation with the donation of the GPUs used in our research.



\bibliography{raffa,ravi,moin}
\bibliographystyle{acl_natbib}



\end{document}

%% file: introduction.tex
\section{Introduction}
\label{sec:introduction}

Most human language understanding is grounded in perception. There is thus growing interest in combining information from language and vision in the NLP and AI communities. 

\begin{figure}
\centering
  \fbox{\includegraphics[height=4cm]{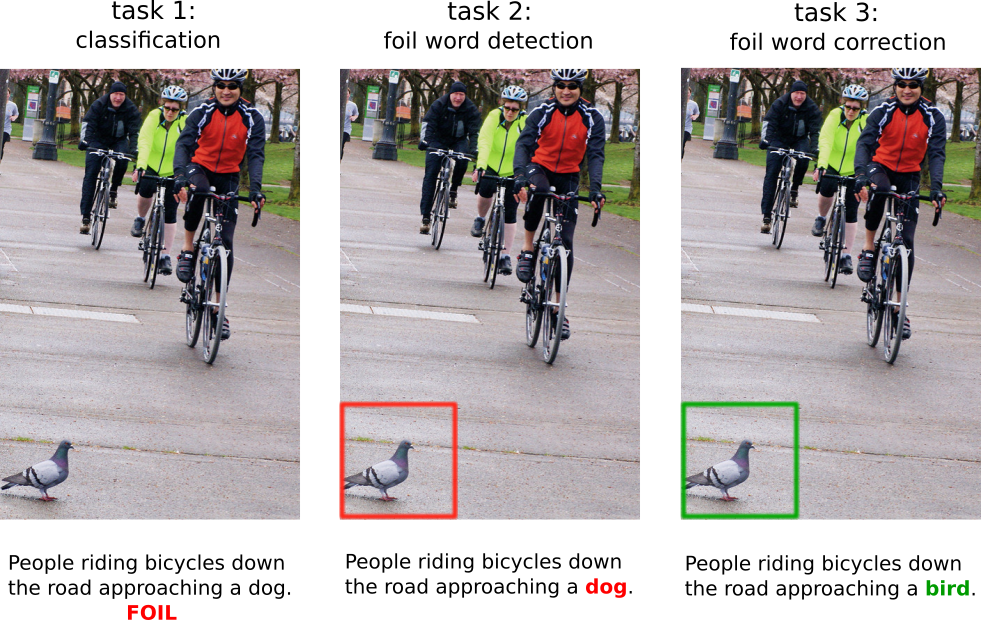}}
  \caption{Is the caption correct or foil (T1)? If it is foil,
    where is the mistake (T2) and which is the word to correct the
    foil one (T3)?}
  \label{fig:problem}
\end{figure}

So far, the primary testbeds of Language and Vision (LaVi) models have been `Visual Question Answering' (VQA)
(e.g.~\citet{anto:vqa15,malinowski2014multi,malinowski2015ask,gao2015you,ren:expl15})
and `Image Captioning' (IC)
(e.g.~\citet{Hodosh:etal:2013,fang2015captions,chen2015mind,donahue2015long,
  karpathy2015deep,vinyals2015show}). Whilst some models have seemed
extremely successful on those tasks, it remains unclear how the
reported results should be interpreted and what those models are
actually learning. There is an emerging feeling in the community
that the VQA task should be revisited, especially as many current dataset can be handled by `blind' models which use language input only, or by simple concatenation of language and vision
features ~\citep{agra:anal16,jabr:revi16,zhang2016yin,goyal2016making}. In
IC too, \citet{hodo:focu16} showed that, contrarily to what prior
research had suggested, the task is far from been solved, since IC
models are not able to distinguish between a correct and incorrect
caption.

Such results indicate that in current datasets, \emph{language provides priors} that make LaVi models successful without truly understanding and integrating language and vision. But problems do not stop at biases. \citet{girs:clev16} also
point out that current data `conflate multiple sources of error, making it hard to pinpoint model weaknesses', thus highlighting the need for \emph{diagnostic datasets}.  Thirdly, existing IC  \emph{evaluation metrics} are sensitive to n-gram overlap and there is a need for measures that better simulate human judgments~\citep{Hodosh:etal:2013,elli:comp14,ande:spic16}.

Our paper tackles the identified issues by proposing an automatic method for creating a large dataset of real images with \textit{minimal language bias} and some \textit{diagnostic} abilities. Our dataset, FOIL (Find One mismatch between Image and Language caption),\footnote{The dataset is available from \url{https://foilunitn.github.io/}} consists of images associated with incorrect captions. The captions are  produced by introducing one single error (or `foil') per caption in existing, human-annotated data (Figure~\ref{fig:problem}). This process results in a challenging error-detection/correction setting  (because the caption is `nearly' correct). It also provides us with a ground truth (we know where the error is) that can be used to objectively measure the performance of current models. 

We propose three tasks based on widely accepted evaluation measures:
we test the ability of the system to a) compute whether a caption is
compatible with the image (T1); b) when it is incompatible, highlight
the mismatch in the caption (T2); c) correct the mistake by replacing
the foil word (T3). 

The dataset presented in this paper (Section~\ref{sec:dataset}) is
built on top of MS-COCO~\citep{lin2014microsoft}, and contains 297,268
datapoints and 97,847 images. We will refer to it as FOIL-COCO.
We evaluate two state-of-the-art VQA models: the popular one by~\citet{anto:vqa15}, and the attention-based model by~\citet{lu:hier16}, and one popular IC model by~\cite{wang2016image}. We show that those models perform close to chance level, while humans can perform the tasks accurately (Section~\ref{sec:experiments}). Section~\ref{sec:analysis} provides an analysis of our results, allowing us to diagnose three failures of LaVi models. First, their coarse representations of language and visual input do not encode suitably structured information to spot mismatches between an utterance and the corresponding scene (tested by T1). Second, their language representation is not fine-grained enough to identify the part of an utterance that causes a mismatch with the image as it is (T2). Third, their visual representation is also too poor to spot and name the visual area that corresponds to a captioning error (T3).

%% file: related_work.tex
\section{Related Work}
\label{sec:related_work}

The image captioning (IC) and visual question answering (VQA) tasks are the most relevant to our work. In IC~\citep{fang2015captions,chen2015mind,donahue2015long,karpathy2015deep,vinyals2015show,wang2016image},
the goal is to generate a caption for a given
image, such that it is both semantically and syntactically correct, and properly
describes the content of that image. In VQA \citep{anto:vqa15,malinowski2014multi,malinowski2015ask,gao2015you,ren:expl15},
the system attempts to answer open-ended questions related to the
content of the image. 
There is a wealth of literature on
both tasks, but we only discuss here the ones most related to our work and
refer the reader to the recent surveys
by~\cite{bernardi2016automatic,wu2016visual}.

Despite their success, it remains unclear whether state-of-the-art LaVi models capture vision and language in a truly integrative fashion. We could identify three types of arguments surrounding the high performance of LaVi models:

{\textbf{(i) Triviality of the LaVi tasks: }} Recent work has shown
that LaVi models heavily rely on language
priors~\citep{ren:expl15,agra:anal16,kafle2016visual}. Even simple
correlation and memorisation can result
in good performance, without the underlying models truly
understanding visual content~\citep{zhou2015simple,jabr:revi16,hodo:focu16}. \citet{zhang2016yin} first unveiled
that there exists a huge bias in the popular VQA dataset
by~\citet{anto:vqa15}: they showed that almost half of all the
questions in this dataset could be answered correctly by using the
question alone and ignoring the image completely. In the same vein, \citet{zhou2015simple} proposed a simple baseline for the
task of VQA. This baseline simply concatenates the Bag of Words (BoW)
features from the question and Convolutional Neural Networks (CNN) features from the image to predict
the answer. They showed that such a simple method can achieve comparable
performance to complex and deep architectures. \citet{jabr:revi16}
proposed a similar model for the task of multiple choice VQA, and
suggested a cross-dataset generalization scheme as an evaluation
criterion for VQA systems. We complement this research by introducing
three new tasks with different levels of difficulty, on which LaVi
models can be evaluated sequentially.

{\textbf{(ii) Need for diagnostics: }}
 To overcome the bias uncovered in previous datasets, several research groups have started  proposing tasks which involve distinguishing distractors from a ground-truth caption for an image. \citet{zhang2016yin} introduced a binary VQA task along with a dataset composed of sets of similar artificial images, allowing for more precise diagnostics of a system's errors. \citet{goyal2016making} balanced the dataset of \citet{anto:vqa15}, collecting a new set of complementary natural images which are similar to existing items in the original dataset, but result in different answers to a common question. \citet{hodo:focu16} also proposed to evaluate a number of state-of-the-art LaVi algorithms in the presence of distractors. Their evaluation was however limited to a small dataset (namely, Flickr30K~\cite{young2014image}) and the caption generation was based on a hand-crafted scheme using only inter-dataset distractors. 
 
Most related to our paper is the work by
~\citet{ding2016understanding}.\cut{\footnote{Please note that this paper
  was only published on \textit{arxiv} a few weeks ago, at the end of
  December 2016. We have not yet evaluated their system against our
  data.}} Like us, they propose to extend the MS-COCO dataset by
generating decoys from human-created image captions. They also suggest
an evaluation  apparently similar to our T1, requiring the LaVi system
to detect the true target caption amongst the decoys. Our efforts,
however, differ in some substantial ways. First, their technique to
create incorrect captions (using BLEU to set an upper similarity
threshold) is so that many of those captions will differ from the gold
description in more than one respect. For instance, the caption
\textit{two elephants standing next to each other in a grass field} is
associated with the decoy \textit{a herd of giraffes standing next to
  each other in a dirt field} (errors: \textit{herd},
\textit{giraffe}, \textit{dirt}) or with \textit{animals are gathering
  next to each other in a dirt field} (error: \textit{dirt};
infelicities: \textit{animals} and \textit{gathering}, which are both
pragmatically odd). Clearly, the more the caption changes in the
decoy, the easier the task becomes. In contrast, the foil captions we
propose only differ from the gold description by \textit{one} word and
are thus more challenging. Secondly, the automatic caption
generation of Ding et al means that `correct' descriptions can be produced,
resulting in some confusion in human responses to the task. We made
sure to prevent such cases, and human performance on our dataset is
thus close to 100\%. We note as well that our task does not require
any complex instructions for the annotation, indicating that it is
intuitive to human beings (see \S\ref{sec:experiments}). Thirdly,
their evaluation is a multiple-choice task, where the system has to
compare all captions to understand which one is \emph{closest} to the
image. This is arguably a simpler task than the one we propose, where
a caption is given and the system is asked to classify it as correct or foil: as we show in \S\ref{sec:experiments}, detecting a \textit{correct} caption is much easier than detecting foils. So evaluating precision on both gold and foil items is crucial.

Finally, \cite{girs:clev16} proposed CLEVR, a dataset for the diagnostic evaluation of VQA systems. This dataset was designed with the explicit goal of enabling detailed analysis of different aspects of visual reasoning, by minimising dataset biases and providing rich ground-truth representations for both images and questions.

{\textbf{(iii) Lack of objective evaluation metrics:}} The evaluation
of Natural Language Generation (NLG) systems is
known to be a hard problem. It is further unclear whether the quality
of LaVi models should be measured using metrics designed for
language-only tasks.
\citet{elli:comp14} performed a sentence-level correlation analysis of NLG evaluation measures against expert human judgements in the context of IC. Their study revealed that most of those metrics were only weakly correlated with human judgements. In the same line of research, \citet{ande:spic16} showed that the most widely-used metrics for IC fail to capture semantic propositional content, which is an essential component of human caption evaluation. They proposed a semantic evaluation metric called SPICE, that measures how effectively image captions recover objects, attributes and the relations between them. In this paper, we tackle this problem by proposing tasks which can be evaluated based on objective metrics for classification/detection error.

%% file: datasets.tex
\section{Dataset}
\label{sec:dataset}
In this section, we describe how we automatically generate FOIL-COCO
datapoints, i.e. image, original and foil caption triples.  We used
the training and validation Microsoft's Common Objects in Context
(MS-COCO) dataset~\cite{lin2014microsoft} (2014 version) as our
starting point. In MS-COCO, each image is described by at least five
descriptions written by humans via Amazon Mechanical Turk (AMT). The
images contains 91 common object categories (e.g. {\em dog, elephant,
  bird, \ldots} and {\em car, bicycle, airplane, \ldots}), from 11
supercategories ({\em Animal}, {\em Vehicle}, resp.), with 82 of them
having more than 5K labeled instances. In total there are 123,287
images with captions (82,783 for training and 40,504 for validation).\footnote{The MS-COCO test set is not available for download.}


\begin{footnotesize}
\begin{table*}[t]
\begin{center}
\begin{tabular}{l|c|c|c|c|}
     & nr. of datapoints & nr. unique images & nr. of tot. captions &
                                                                      nr. target::foil
                                                                      pairs
  \\\hline                          Train & 197,788 & 65,697 &  395,576  & 256 \\
  Test & 99,480 & 32,150   & 198,960  & 216 \\
\end{tabular}
\caption{Composition of FOIL-COCO.}\label{tab:dataset}
\end{center}
\end{table*}
\end{footnotesize}

\begin{figure*}[t]
\centering
  \fbox{\includegraphics[width=15cm, height=6cm]{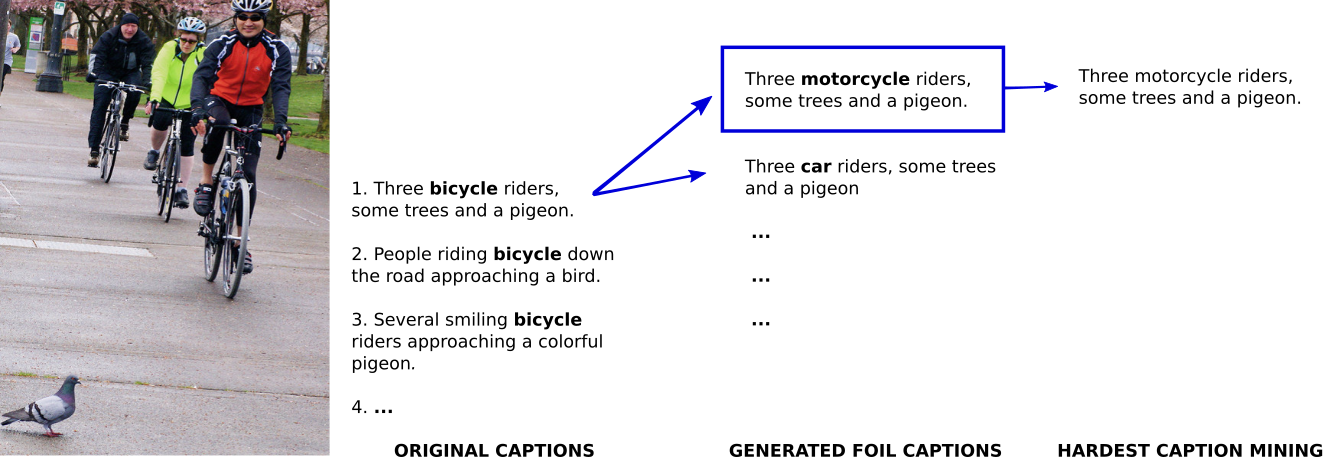}}
  \caption{The main aspects of the foil caption generation process.
Left column: some of the original COCO captions associated with an image. In bold we highlight one of the target words (bicycle), chosen because it is mentioned by more than one annotator.
Middle column: For each original caption and each chosen target word, different foil captions are generated by replacing the target word with all  possible candidate foil replacements.
Right column: A single caption is selected amongst all foil
candidates. We select the `hardest' caption, according to 
Neuraltalk model, trained using only the original 
captions.}\label{fig:process}
\end{figure*}

Our data generation process consists of four main steps, as described below. The last two steps are
illustrated in Figure~\ref{fig:process}.

\textbf{1. Generation of replacement word pairs} We want to replace one noun in the original caption (the \textit{target}) with an incorrect but similar word (the \emph{foil}). To do this, we take the labels of MS-COCO
categories, and we pair together words belonging to the same
supercategory (e.g., bicycle::motorcycle, bicycle::car, bird::dog). We use as our vocabulary
73 out of the 91 MS-COCO categories, leaving out those categories
that are multi-word expressions (e.g. traffic light). We
thus obtain 472 target::foil pairs.

\textbf{2. Splitting of replacement pairs into training and
  testing} To avoid the models learning trivial correlations due
to replacement frequency, we randomly split, within each
supercategory, the candidate target::foil pairs which are used to
generate the captions of the training vs.\ test sets.  We
obtain 256 pairs, built out of 72 target and 70 foil words, for the training set, and 216 pairs, containing
73 target and 71 foil words, for the test set.

\textbf{3. Generation of foil captions}
 We would like to generate foil captions by replacing only target words
which refer to {\em visually salient objects}.  To this end, given an
image, we replace only those target words that occur in more than one
MS-COCO caption associated with that image. Moreover, we want to use
foils which are {\em not visually present}, i.e. that refer to visual
content not present in the image. Hence, given an image, we only replace a word with foils that
are not among the labels (objects) annotated in MS-COCO for that
image. 
We use the images from the MS-COCO training and validation
sets to generate our training and test sets, respectively. We obtain
2,229,899 for training and 1,097,012 captions for testing.

\textbf{4. Mining the hardest foil caption for each image} 
To eliminate possible visual-language dataset bias, out of all foil captions generated in step
3, we select only the hardest one.  For this purpose, we need to
model the visual-language bias of the dataset. To this end,
we use Neuraltalk\footnote{\url{https://github.com/karpathy/neuraltalk}} 
\cite{karpathy2015deep}, one of the state-of-the-art image
captioning systems, pre-trained on MS-COCO.   Neuraltalk is based on an LSTM
  which takes as input an image and generates a sentence describing
  its content.  We
  obtain a neural network ${\cal N}$ that implicitly represents the
  visual-language bias through its weights.  We use ${\cal N}$ to
  approximate the conditional probability of a caption $C$ given a
  dataset $T$  and 
and an image $I$ ($P(C|I,T)$). This is obtained by simply using the  loss
$l(C,{\cal N}(I)) $ i.e., the error obtained by  comparing the
pseudo-ground truth $C$ with the sentence predicted by  ${\cal N}$:
$P(C|I,T) = 1 - l(C,{\cal N}(I))$ (we refer to  \cite{karpathy2015deep}
for more details on how $l()$ is computed). $P(C|I,T)$ is used to select the hardest foil among all
the possible foil captions, i.e. the one with the highest  probability according to the dataset bias
learned by ${\cal N}$. 
Through this process, we obtain
197,788  and 99,480 original::foil caption
  pairs for the training and test sets, respectively. None of the 
  target::foil word pairs are filtered out by this mining
  process.

The final FOIL-COCO dataset consists of 297,268 datapoints (197,788 in
training and 99,480 in test set). All the 11 MS-COCO
supercategories are represented in our dataset and contain 73 categories from
the 91 MS-COCO ones (4.8 categories per supercategory on
average.) Further details are reported in Table~\ref{tab:dataset}.

%% file: experiments.tex
\section{Experiments and Results}
\label{sec:experiments}

We conduct three tasks, as presented below:
\vspace{1mm}

{\textbf{Task 1 (T1): Correct vs.\ foil classification}}
Given an image and a caption, the model is asked to mark whether the caption
is correct or wrong. The aim is to understand whether LaVi models can spot mismatches between their coarse representations of language and visual input.
\vspace{1mm} 

{\textbf{Task 2 (T2): Foil word detection}} Given an image and a
  foil caption, the model has to detect the foil word. The aim is to evaluate the understanding of the system at the word level. 
  In order to systematically check the system's performance with different
  prior information, we test two different
  settings: the foil has to be selected amongst (a) only the nouns or (b) all content words in the caption.
\vspace{1mm}

{\textbf{Task 3 (T3): Foil word correction}} Given an image, a
  foil caption and the foil word, the model has to detect the foil
  and provide its correction. The aim is to check whether the system's visual representation is fine-grained enough to be able to extract the information necessary to correct the error.
  For efficiency reasons, we operationalise this task by asking
  models to select a correction from the set of target words, rather than the whole dataset vocabulary
  (viz.\ more than 10K words).

\subsection{Models}
\label{sec:models}

We evaluate both VQA and IC models against our tasks.  For the
former, we use two of the three models evaluated
in~\cite{goyal2016making} against a balanced VQA
dataset\cut{~\footnote{We have not evaluated the Multimodal Compact
    Bilinear Pooling~\cite{fuki:mult16} against our tasks, however
    in~\cite{goyal2016making} this model accuracy is higher than the
    HieCoAtt of 4.3\% -- which in the context of our results is a
    rather small difference.}}. For the latter, we use the
multimodal bi-directional LSTM, proposed in~\cite{wang2016image}, and
adapted for our tasks.

\paragraph{LSTM + norm I:} We use the best
performing VQA model in~\cite{anto:vqa15} (deeper LSTM  + norm I).
This model uses a two stack Long-Short Term
Memory (LSTM) to encode the questions and the last fully connected
layer of VGGNet to encode images. Both image embedding and
caption embedding are projected into a 1024-dimensional feature
space. Following~\cite{anto:vqa15}, we have normalised the image
feature before projecting it. The combination
of these two projected embeddings is performed by a point-wise
multiplication. The multi-model representation thus obtained is used for
the classification, which is performed by a multi-layer perceptron (MLP) classifier.

\paragraph{HieCoAtt:} We use the Hierarchical Co-Attention model proposed by~\cite{lu:hier16}
that co-attends to both the image and the question to solve the
task. In particular, we evaluate the `alternate' version, i.e. the
model that sequentially alternates between generating some attention over the
image and question. It does so in a hierarchical way by
starting from the word-level, then going to the phrase and then to the
entire sentence-level. These levels are combined recursively to
produce the distribution over the foil vs. correct captions.

\paragraph{IC-Wang:} Amongst the IC models, we choose the multimodal
bi-directional LSTM (Bi-LSTM) model proposed
in~\cite{wang2016image}. This model predicts a word in a sentence by
considering both the past and future context, as sentences are
fed to the LSTM in forward and backward order. The model consists of
three modules: a CNN for encoding image inputs, a Text-LSTM (T-LSTM)
for encoding sentence inputs, a Multimodal LSTM (M-LSTM) for embedding
visual and textual vectors to a common semantic space and decoding to
sentence. The bidirectional LSTM is implemented with two separate LSTM
layers.

\paragraph{Baselines:} We compare the SoA models above against two
baselines. For the classification task, we use a \textbf{Blind} LSTM
model followed by a fully connected layer and softmax and train it
only on captions as input to predict the answer. In addition, we
evaluate the \textbf{CNN+LSTM} model, where visual and textual
features are simply concatenated.

\paragraph{The models at work on our three tasks}
For the \emph{classification task} (T1), the baselines and VQA models
can be applied directly. We adapt the generative IC model to
perform the classification task as follows.  Given a test image $I$
and a test caption, for each word $w_t$ in the test caption, we 
remove the word and use the model to generate new captions in which
the $w_t$ has been replaced by the word $v_t$ predicted by the model
($w_{1}$,...,$w_{t-1}, v_t, w_{t-1}$,...,$w_{n}$).  We then compare
the conditional probability of the test caption with all the captions
generated from it by replacing $w_t$ with $v_t$.  When all the
conditional probabilities of the generated captions are lower than the
one assigned to the test caption the latter is classified as good,
otherwise as foil. For the other tasks, the models have been trained
on T1. To perform the \emph{foil word detection task} (T2), for the
VQA models, we apply the occlusion
method. Following~\cite{goya:towa16}, we systematically occlude
subsets of the language input, forward propagate the masked input
through the model, and compute the change in the probability of the
answer predicted with the unmasked original input. For the IC model,
similarly to T1, we sequentially generate new captions from the foil
one by replacing, one by one, the words in it and computing the
conditional probability of the foil caption and the one generated from
it. The word whose replacement generate the caption with the highest
conditional probabilities is taken to be the foil word.  Finally, to
evaluate the models on the \emph{error correction task} (T3), we
apply the linear regression method over all the target words and
select the target word which has the highest probability of making
that wrong caption correct with respect to the given image.

\cut{For the classification task, we used a
\textbf{Blind} model, viz. MLP+BoW~\cite{jabr:revi16}. This model only
accepts captions as input to predict the answer. Differently
from~\cite{jabr:revi16} instead of concatenating BoW representations
of caption and corresponding answer as input to MLP, we only input BoW
representation of caption to the MLP network and use it as
classifier. Apart from that, we have used the same number of
parameters, viz.\ MLP has 2 hidden layers having 8,192 hidden units
and dropout is used after the first layer. For the error detection and
error classification tasks, we compare the models' results against
chance.\footnote{The average number of nouns per caption is 4.3 and
  average number of content words (viz., after removing the stop
  words) is 6.3.  Similary for T3, there are 72 possible target words
  for a given foil word.}}

\paragraph{Upper-bound} Using Crowdflower, we collected human answers
from 738 native English speakers for 984 image-caption pairs randomly
selected from the test set. Subjects were given an image and a caption
and had to decide whether it was correct or wrong (T1). If they
thought it was wrong, they were required to mark the error in the
caption (T2). We collected 2952 judgements (i.e. 3 judgements per pair
and 4 judgements per rater) and computed human accuracy in T1 when
considering as answer (a) the one provided by at least 2 out of 3
annotators (\textit{majority}) and (b) the one provided by all 3
annotators (\textit{unanimity}). The same procedure was adopted for
computing accuracies in T2. Accuracies in both T1 an T2 are reported
in Table~\ref{tab:results}. As can be seen, in the \textit{majority}
setting annotators are quasi-perfect in classifying captions (92.89\%)
and detecting foil words (97.00\%). Though lower, accuracies in the
\textit{unanimity} setting are still very high, with raters providing
the correct answer in 3 out of 4 cases in both tasks.
Hence, although we have collected human answers only on a
rather small subset of the test set, we believe their results are
representative of how easy the tasks are for humans.

\subsection{Results}
\label{sec:results}

As shown in Table~\ref{tab:results}, the FOIL-COCO dataset is
challenging. On T1, for which the chance level is $50.00\%$, the
`blind', language-only model, does badly with an accuracy of $55.62\%$
($25.04\%$ on foil captions), demonstrating that language bias is
minimal. By adding visual information, CNN+LSTM, the overall accuracy
increases by $5.45 \%$ ($7.94\%$ on foil captions.)  reaching
$61.07\%$ (resp. $32.98\%$).  Both SoA VQA and IC models do
significantly worse than humans on both T1 and T2. The VQA systems
show a strong bias towards correct captions and poor overall
performance. They only identify $34.51\%$ (LSTM +norm I) and $36.38\%$
(HieCoAtt) of the incorrect captions (T1). On the other hand, the IC
model tends to be biased toward the foil captions, on which it
achieves an accuracy of $45.44\%$, higher than the VQA models. But the
overall accuracy ($42.21\%$) is poorer than the one obtained by the
two baselines. On the foil word detection task, when considering only
nouns as possible foil word, both the IC and the LSTM+norm I models
perform close to chance level, and the HieCoAtt performs somewhat
better, reaching $38.79\%$. Similar results are obtained when
considering all words in the caption as possible foil. Finally, the
VQA models' accuracy on foil word correction (T3) is extremely low, at
$4.7\%$ (LSTM +norm I) and $4.21\%$ (HieCoAtt). The result on T3 makes
it clear that the VQA systems are unable to extract from the image
representation the information needed to correct the foil: despite
being told which element in the caption is wrong, they are not able to
zoom into the correct part of the image to provide a correction, or if
they are, cannot name the object in that region. The IC model performs
better compared to the other models, having an accuracy that is
20,78\% higher than chance level.

\begin{table}[ht]

\begin{tabular}{|l|c|c|c|}
\hline
 \multicolumn{4}{|c|}{ \textbf{T1:} Classification task}\\\hline    
                 & Overall  & Correct & Foil \\\hline 
\cut{Blind BoW & 43.95 & 68.36 & 19.53 \\\hline}
Blind  & 55.62 & 86.20 & 25.04 \\
CNN+LSTM  & 61.07 & 89.16 & 32.98 \\ \hline
IC-Wang  & 42.21 & 38.98 & 45.44 \\\hline
\cut{CNN+BoW  & 60.62 & 88.67 & 32.57 \\}

LSTM + norm I  &  63.26 & \textbf{92.02} & 34.51\\
HieCoAtt            & \textbf{64.14} & 91.89 & \textbf{36.38} \\\hline
Human (\textit{majority})            & 92.89  & 91.24 & 94.52 \\
Human (\textit{unanimity})            & 76.32  & 73.73 & 78.90 \\
\hline
\end{tabular}

\vspace{3mm}
\begin{tabular}{|l|c|c|}
\hline
\multicolumn{3}{|c|}{\textbf{T2:} Foil word detection task} \\\hline
           &  nouns & all content words\\\hline
Chance & 23.25 & 15.87    \\\hline
IC-Wang & 27.59  &  23.32  \\\hline
LSTM + norm I  & 26.32  &  24.25  \\
HieCoAtt & \textbf{38.79} & \textbf{33.69}  \\\hline
Human (\textit{majority}) & &   97.00\\
Human (\textit{unanimity}) & &   73.60\\
\hline
\end{tabular}
\vspace{3mm}

\begin{center}

\begin{tabular}{|l|c|}
\hline
\multicolumn{2}{|c|}{\textbf{T3:} Foil word correction task}\\\hline
             &   all target words\\\hline
Chance &  1.38 \\\hline
IC-Wang  & \textbf{22.16} \\\hline
LSTM + norm I  &  4.7\\
HieCoAtt &   4.21  \\
\hline
\end{tabular}
\end{center}

\caption{\textbf{T1:} Accuracy for the \emph{classification} task, relatively to  all image-caption pairs (overall) and by type of caption (correct  vs.\ foil); \textbf{T2:} Accuracy for the \emph{foil word detection} task, when the foil  is known to be among the nouns only or when it is known to be among  all the content words; \textbf{T3:} Accuracy for the \emph{foil word correction} task when the  correct word has to be chosen  among any of the target words.}\label{tab:results}
\end{table}

%% file: analysis.tex
\section{Analysis}
\label{sec:analysis}

\begin{table*}[ht]
\centering
\begin{tabular}{|c|c|c|c|c|c|c|	}
\hline
Super-category & \makecell{No. of object} & \makecell{No. of foil \\ captions} & \makecell{Acc. using \\ LSTM + norm I} & \makecell{Acc. using \\ HieCoAtt} \\\hline
outdoor &2 &107& 2.80 & 0.93 \\\hline
 food & 9 & 10407 &  22.00 & 26.59  \\\hline
 indoor & 6 & 4911 &  30.74 & 27.97 \\\hline
 appliance & 5 & 2811 &  32.72 & 34.54 \\\hline
 sports & 10 &  16276 & 31.57 & 31.61 \\\hline
 animal & 10 & 21982 &  39.03 & 43.18  \\\hline
 vehicle & 8 & 16514 & 34.38 & 40.09\\\hline
 furniture & 5 & 13625 & 33.27 & 33.13\\\hline
 accessory & 5 & 3040 & 49.53 & 31.80\\\hline
 electronic & 6&  5615 & 45.82 & 43.47\\\hline
 kitchen & 7 & 4192 & 38.19 & 45.34\\\hline
\end{tabular}
\caption{Classification Accuracy of foil captions by Super Categories (T1). The
  No. of the objects and the No. of foil captions refer to the test
  set. The training set has a similar distribution.}
\label{tab:superClass}
\end{table*}

We performed a mixed-effect logistic regression analysis in order to
check whether the behavior of the best performing models in T1,
namely the VQA models, can be predicted by various linguistic
variables. We included: 1) semantic similarity between the original
word and the foil (computed as the cosine between the two
corresponding \texttt{word2vec} embeddings~\cite{mikolov2013efficient}); 2) frequency of original
word in FOIL-COCO captions; 3) frequency of the foil word in FOIL-COCO
captions; 4) length of the caption (number of words). The mixed-effect
model was performed to get rid of possible effects due to either
object supercategory (indoor, food, vehicle, etc.) or target::foil
pair (e.g., zebra::giraffe, boat::airplane, etc.). For both LSTM +
norm I and HieCoAtt, \texttt{word2vec} similarity, frequency of the
original word, and frequency of the foil word turned out to be highly
reliable predictors of the model's response. The higher the values of
these variables, the more the models tend to provide the wrong
output. That is, when the foil word (e.g.\ \textit{cat}) is
semantically very similar to the original one (e.g.\ \textit{dog}),
the models tend to wrongly classify the caption as `correct'. The same
holds for frequency values. In particular, the higher the frequency of
both the original word and the foil one, the more the models
fail. This indicates that systems find it difficult to distinguish
related concepts at the text-vision interface, and also that they may
tend to be biased towards frequently occurring concepts, `seeing them
everywhere' even when they are not present in the image. Caption
length turned out to be only a partially reliable predictor in the
LSTM + norm I model, whereas it is a reliable predictor in
HieCoAtt. In particular, the longer the caption, the harder for the model
to spot that there is a foil word that makes the caption wrong.

As revealed by the fairly high variance explained by the random effect
related to target::foil pairs in the regression analysis, both models
perform very well on some target::foil pairs, but fail on some
others (see leftmost part of Table~\ref{tab:pairs} for same examples of easy/hard
target::foil pairs). Moreover, the variance explained by the random effect
related to object supercategory is reported in Table~\ref{tab:superClass}. As can be seen, for some supercategories accuracies are significatively higher than for others (compare, e.g., `electronic' and `outdoor').

\cut{
\begin{figure}
\begin{footnotesize}
\begin{verbatim}
              Estimate Std. Error     
capt_length  5.881e-03  3.135e-03  .  
Word2vec    -8.924e+00  2.171e-01  ***
Target_Freq -3.045e+01  9.713e-01  ***
Foil_Freq   -1.873e+02  1.133e+00  ***
\end{verbatim}
\end{footnotesize}
\caption{Regression analysis: models' accuracy on Task 1. $***$ indicates significance at $p < 0.001$}
\end{figure}\label{fig:regression}}

\begin{table*}
\begin{center}
\begin{footnotesize}
\begin{tabular}{cc}
\hspace*{-2mm} \begin{tabular}{l|r||l|r}
\multicolumn{2}{c}{Top-5} & \multicolumn{2}{c}{Bottom-5}\\\hline
\multicolumn{4}{c}{T1: LSTM + norm I}\\\hline
racket::glove &  100 & motorcycle::airplane & 0\\
racket::kite &  97.29 & bicycle::airplane &  0 \\
couch::toilet &  97.11 & drier::scissors &  0\\
racket::skis &  95.23  & bus::airplane &  0.35\\
giraffe::sheep &  95.09 & zebra::giraffe &  0.43\\\hline
\multicolumn{4}{c}{T1: HieCoAtt}\\\hline
tie::handbag &  100 & drier::scissors  &  0 \\
snowboard::glove &  100  & fork::glass  &  0 \\
racket::skis & 100 & handbag::tie  &  0\\
racket::glove &  100 & motorcycle::airplane  & 0\\
backpack::handbag & 100  & train::airplane  & 0\\
\end{tabular}

\hspace*{-2mm} & \hspace*{-2mm}
\begin{tabular}{l|r||l|r}
\multicolumn{2}{c}{Top-5} & \multicolumn{2}{c}{Bottom-5}\\\hline
\multicolumn{4}{c}{T2: LSTM + norm I}\\\hline
drier::scissors & 100 & glove::skis &  0\\
zebra::giraffe &  88.98 & snowboard::racket  & 0   \\
boat::airplane &  87.87 & donut::apple &   0\\
truck::airplane & 85.71 & glove::surfboard &  0 \\
train::airplane &  81.93 & spoon::bottle &  0  \\\hline
\multicolumn{4}{c}{T2: HieCoAtt}\\\hline
zebra::elephant &  94.92 & drier::scissors &  0\\
backpack::handbag &  94.44 & handbag::tie &  0\\
cow::zebra &  93.33 & broccoli:orange &  1.47 \\
bird::sheep &  93.11 & zebra::giraffe &  1.96 \\
orange::carrot & 92.37  & boat::airplane &  2.09 \\
\end{tabular}
\end{tabular}
\caption{Easiest and hardest target::foil pairs: T1 (caption
  classification) and T2 (foil word detection).}\label{tab:pairs}
\end{footnotesize}
\end{center}
\end{table*}

In a separate analysis, we also checked whether there was any correlation between results
and the position of the foil in the sentence, to ensure the models did
not profit from any undesirable artifacts of the data. We did not find any such
correlation.

To better understand results on T2, we performed an analysis
investigating the performance of the VQA models on different
target::foil pairs. As reported in Table~\ref{tab:pairs} (right), both
models perform nearly perfectly with some pairs and very badly with
others. At first glance, it can be noticed that LSTM + norm I is very
effective with pairs involving vehicles (\textit{airplane},
\textit{truck}, etc.), whereas HieCoAtt seems more effective with
pairs involving animate nouns (i.e. animals), though more in depth
analysis is needed on this point. More interestingly, some pairs that
are found to be predicted almost perfectly by LSTM + I norm, namely
boat::airplane, zebra::giraffe, and drier::scissors, turn out to be
among the Bottom-5 cases in HieCoAtt. This suggests, on the one hand,
that the two VQA models use different strategies to perform the
task. On the other hand, it shows that our dataset does not contain
cases that are a priori easy for any model.

The results of IC-Wang on T3 are much higher than LSTM + norm I and HieCoAtt, although it is outperformed by or is on par with HieCoAtton on T1-T2. Our interpretation is that this behaviour is related to the discriminative/generative nature of our tasks. Specifically, T1 and T2 are discriminative tasks and LSTM + norm I and HieCoAtt are discriminative models. Conversely, T3 is a generative task (a word needs to be generated) and IC-Wang is a generative model. It would be interesting to test other IC models on T3 and compare their results against the ones reported here. However, note that IC-Wang is `tailored' for T3 because it takes as input the whole sentence (minus the word to be generated), while common sequential IC approaches can only generate a word depending on the previous words in the sentence.

\begin{figure}
\centering
  \fbox{\includegraphics[height=4cm]{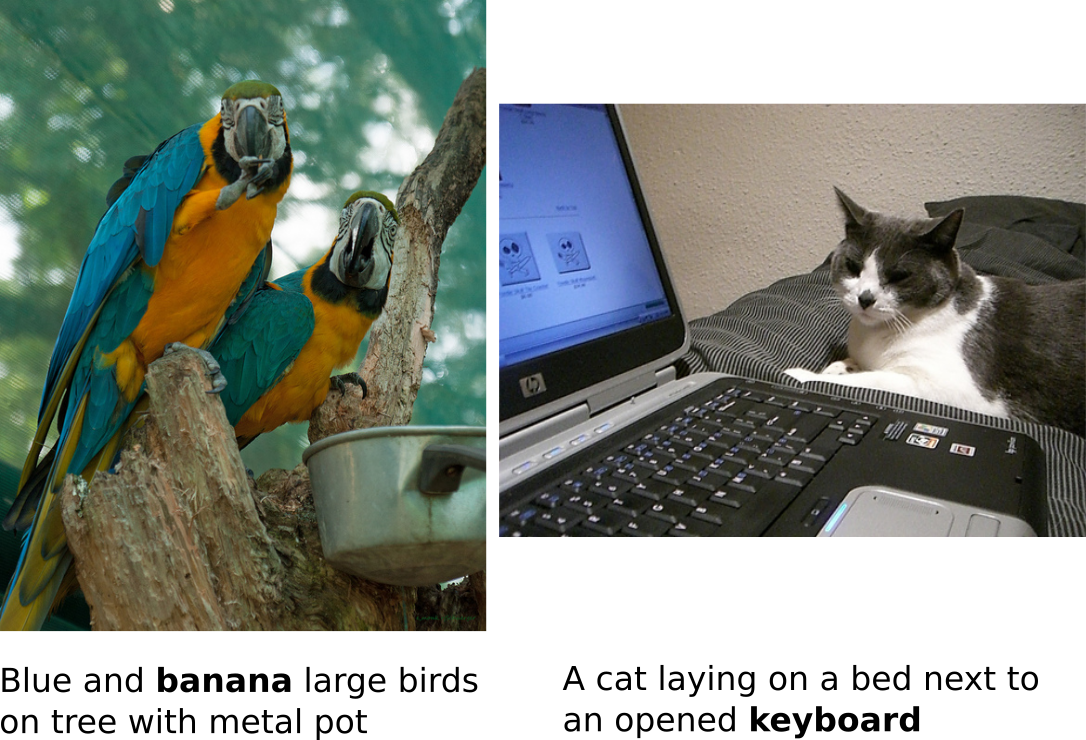}}
  \caption{Two cases of foil image-caption pairs that are judged as correct by all annotators.}
  \label{fig:qual}
\end{figure}

As far as human performance is concerned, both T1 and T2 turn out to be extremely easy. In T1, image-caption pairs were correctly judged as correct/wrong in overall 914 out of 984 cases (92.89\%) in the \textit{majority} setting. In the \textit{unanimity} setting, the correct response was provided in 751 out of 984 cases (76.32\%). Judging foil captions turns out to be slightly easier than judging correct captions in both settings, probably due to the presence of typos and misspellings that sometimes occur in the original caption (e.g. raters judge as wrong the original caption \textit{People playing ball with a drown and white dog}, where `brown' was misspelled as `drown'). To better understand which factors contribute to make the task harder, we qualitatively analyse those cases where all annotators provided a wrong judgement for an image-caption pair. As partly expected, almost all cases where original captions (thus correct for the given image) are judged as being wrong are cases where the original caption is indeed incorrect. For example, a caption using the word `motorcycle' to refer to a bicycle in the image is judged as wrong. More interesting are those cases where all raters agreed in considering as correct image-caption pairs that are instead foil. Here, it seems that vagueness as well as certain metaphorical properties of language are at play: human annotators judged as correct a caption describing \textit{Blue and banana large birds on tree with metal pot} (see Fig~\ref{fig:qual}, left), where `banana' replaced `orange'. Similarly, all raters judged as correct the caption \textit{A cat laying on a bed next to an opened keyboard} (see Fig~\ref{fig:qual}, right), where the cat is instead laying next to an opened laptop.

Focusing on T2, it is interesting to report that among the correctly-classified foil cases, annotators provided the target word in 97\% and 73.6\% of cases in the \textit{majority} and \textit{unanimity} setting, respectively. This further indicates that finding the foil word in the caption is a rather trivial task for humans.

%% file: conclusions.tex
\section{Conclusion}
\label{sec:conclusions}

We have introduced FOIL-COCO, a large dataset of images associated with both correct and foil captions. The error production is automatically generated, but carefully thought out, making the task of spotting foils particularly challenging. By associating the dataset with a series of tasks, we allow for diagnosing various failures of current LaVi systems, from their coarse understanding of the correspondence between text and vision to their grasp of language and image structure.

Our hypothesis is that systems which, like humans, deeply integrate
the language and vision modalities, should spot foil captions quite
easily. The SoA LaVi models we have tested fall through
that test, implying that they fail to integrate the two modalities. To
complete the analysis of these results, we plan to carry out a further
task, namely ask the system to detect in the image the area that
produces the mismatch with the foil word (the red box around the
bird in Figure~\ref{fig:problem}.) This extra step would allow us to
fully diagnose the failure of the tested systems and confirm what is
implicit in our results from task 3: that the algorithms are unable to
map particular elements of the text to their visual counterparts. We
note that the addition of this extra step will move this work closer
to the textual/visual explanation research (e.g.,
\citep{park:atte16,selv:grad16}). We will then have a pipeline able to
not only test whether a mistake can be detected, but also whether the
system can explain its decision: `the wrong word is \textit{dog}
because the cyclists are in fact approaching a bird, there, in the
image'.

LaVi models are a great success of recent research, and we are impressed by the amount of ideas, data and models produced in this stimulating area. With our work, we would like to push the community to think of ways that models can better merge language and vision modalites, instead of merely using one to supplement the other.